\documentclass[10pt,twocolumn,letterpaper]{article}

\usepackage{iccv}              

\definecolor{iccvblue}{rgb}{0.21,0.49,0.74}
\usepackage[pagebackref,breaklinks,colorlinks,allcolors=iccvblue]{hyperref}

\usepackage{pifont}
\usepackage{amssymb}
\newcommand{\xmark}{\ding{55}}
\usepackage{booktabs}
\usepackage{multirow} 
\usepackage{graphicx} 
\usepackage{adjustbox}
\definecolor{mymod}{RGB}{0,0,255}
\definecolor{mytodo}{RGB}{255,0,0}
\definecolor{mydel}{RGB}{127,127,127}

\def\paperID{7435} 
\def\confName{ICCV}
\def\confYear{2025}

\title{Height-Fidelity Dense Global Fusion for Multi-modal 3D Object Detection\vspace{-0.cm}}

\author{Hanshi Wang$^{1,2,3,5}$\thanks{This work was completed during Hanshi's remote internship at SJTU.}, Jin Gao$^{1,2}$, Weiming Hu$^{1,2,5,6}$, Zhipeng Zhang$^{3,4}$\thanks{Corresponding author.}\\
$^1$State Key Laboratory of Multimodal Artificial Intelligence Systems (MAIS), CASIA\\
$^2$School of Artificial Intelligence, University of Chinese Academy of Sciences \\
$^3$School of Artificial Intelligence, Shanghai Jiao Tong University $^4$Anyverse Intelligence\\
$^5$Beijing Key Laboratory of Super Intelligent Security of Multi-Modal 
Information\\
$^6$School of Information Science and Technology, ShanghaiTech University\\
{\tt\small \{hanshi.wang.cv, zhipeng.zhang.cv\}@outlook.com, \{wmhu, jin.gao\}@nlpr.ia.ac.cn}
}

\begin{document}
\maketitle
\begin{abstract}

We present the first work demonstrating that a \textbf{pure Mamba} block can achieve \textbf{efficient Dense Global Fusion}, meanwhile guaranteeing top performance for camera-LiDAR multi-modal 3D object detection.
Our motivation stems from the observation that existing fusion strategies are constrained by their inability to simultaneously achieve efficiency, long-range modeling, and retaining complete scene information. 
Inspired by recent advances in state-space models (SSMs)~\cite{gu2023mamba} and linear attention~\cite{sun2023retentive,peng2023rwkv}, we leverage their linear complexity and long-range modeling capabilities to address these challenges.
However, this is non-trivial since our experiments reveal that simply adopting efficient linear-complexity methods does not necessarily yield improvements and may even degrade performance. We attribute this degradation to the loss of height information during multi-modal alignment, leading to deviations in sequence order.
To resolve this, we propose height-fidelity LiDAR encoding that preserves precise height information through voxel compression in continuous space, thereby enhancing camera-LiDAR alignment. Subsequently, we introduce the Hybrid Mamba Block, which leverages the enriched height-informed features to conduct local and global contextual learning. By integrating these components, our method achieves state-of-the-art performance with the top-tire NDS score of 75.0 on the nuScenes~\cite{caesar2020nuscenes} validation benchmark, even surpassing methods that utilize high-resolution inputs. Meanwhile, our method maintains efficiency, achieving faster inference speed than most recent state-of-the-art methods. 
Code is available at \url{https://github.com/AutoLab-SAI-SJTU/MambaFusion}
\end{abstract}

\section{Introduction}

\begin{figure}[t]
\vspace{-0.5cm}
    \centering
    \footnotesize
    \includegraphics[width=0.98\linewidth]{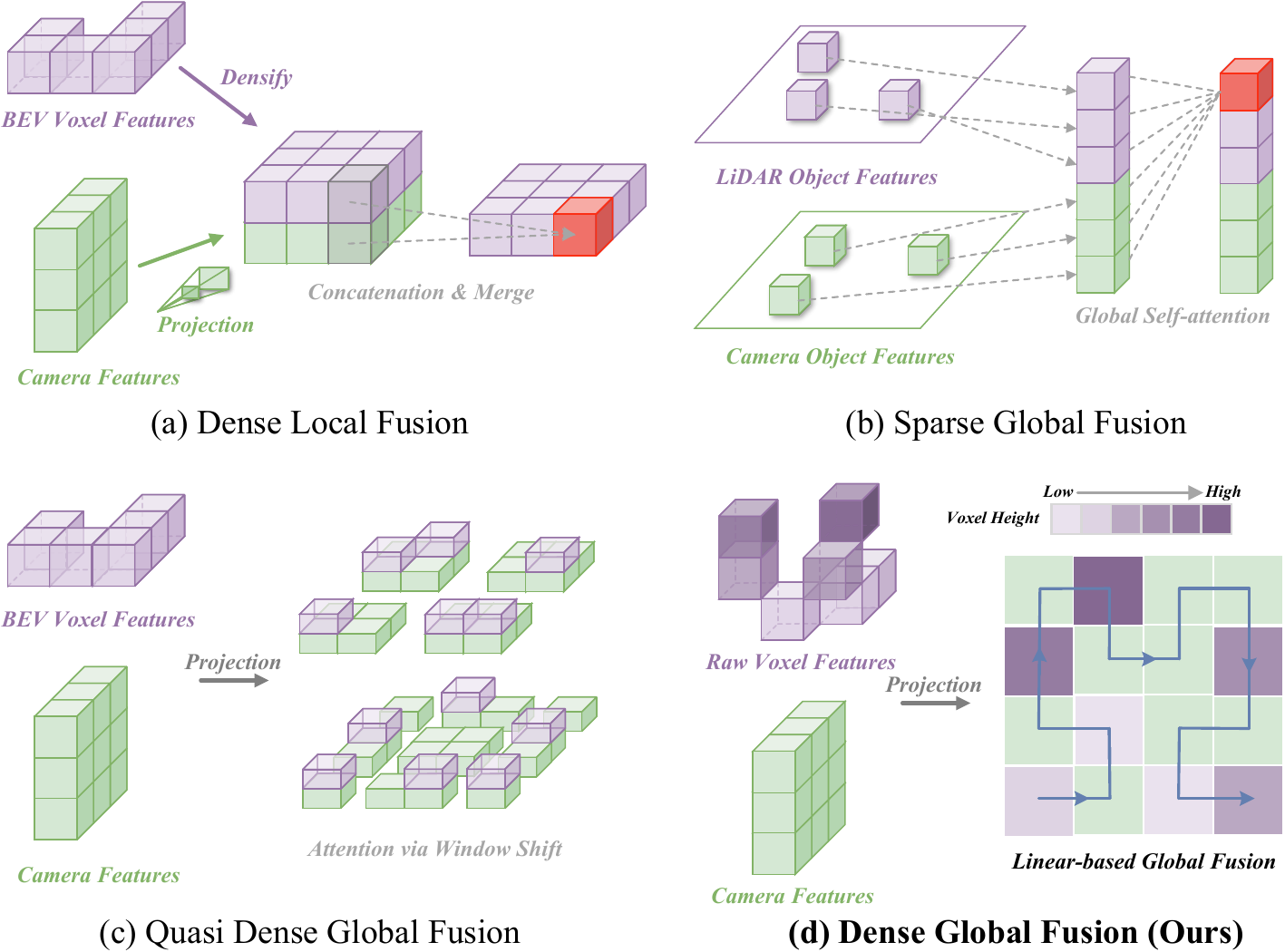}
    \vspace{-0.1cm}
    \caption{Illustration of the LiDAR and camera fusion framework: a) Dense Local Fusion, b) Sparse Global Fusion, c) Quasi Dense Global Fusion, \textbf{d) Dense Global Fusion {(Ours)}. }}
    \label{fig:intro}
    \vspace{-0.6cm}
\end{figure}

3D object detection is a foundational task in autonomous driving, enabling vehicles to accurately interpret their surroundings. As the complexity of driving environments escalates, the demand for robust detection methods becomes increasingly critical. LiDAR and camera are two commonly employed sensors in autonomous vehicles, each possessing distinct advantages. Specifically, LiDAR excels in delivering precise spatial and depth information, while cameras provide rich visual textures and color details. Their complementary nature offers a valuable opportunity for multi-sensor fusion, which can enhance detection accuracy and robustness across diverse and challenging scenarios.

Fig.~\ref{fig:intro} (a), (b), and (c) summarize three types of camera-LiDAR fusion strategies present in existing frameworks, categorized by the range of context utilized in representation learning. 
More specifically, Dense Local Fusion~\cite{liu2023bevfusion,liang2022bevfusion} (Fig.~\ref{fig:intro}(a)) efficiently integrates LiDAR and visual features from aligned local fields in Bird's Eye View (BEV) space. Despite the simplicity, its capability is constrained by the lack of comprehensive contextual information from the surrounding environment (\textbf{efficient but short-sighted}). To leverage long-range dependencies, Sparse Global Fusion~\cite{zhou2023sparsefusion,zhang2024sparselif} (Fig.~\ref{fig:intro}(b)) employs self-attention to facilitate interaction among all object features extracted from the predicted bounding boxes of both modalities. While this approach effectively models long-range global dependency, it inadvertently discards scene information that, although not belonging to predefined categories, is crucial for accurately localizing targets (\textbf{efficient and far-sighted, but un-omniscient\footnote{“Omniscient”: each position can interact with all scene information.}}). Quasi Dense Global Fusion~\cite{wang2023unitr,zhu2024flatfusion} (Fig.~\ref{fig:intro}(c)) represents a compromise between the aforementioned frameworks, drawing inspiration from Swin Transformer~\cite{vaswani2017attention} to shift the local fused BEV window. 
However, the inability to directly interact with global features hinders comprehensive scene understanding (\textbf{efficient but short-sighted and un-omniscient}). 
We summarize the characteristics of these frameworks in Tab.~\ref{tab:categorization}. Consequently, a critical question naturally arises: \textit{Is an efficient, far-sighted, and omniscient long-range fusion framework possible?}

Before answering the question, we first discuss one key requirement for camera-LiDAR global fusion: efficiency. Efficiency is crucial, as marginal performance improvements that incur substantial computational costs are nearly insignificant for autonomous driving applications. This rationale leads us to forgo the straightforward approach of applying self-attention to the multimodal features, given the quadratic complexity associated with this operator. 
Fortunately, current linear complexity methods (\textit{e.g.} Mamba and linear attention) may offer a solution. However, the adaption is non-trivial. For instance, when we substitute the fusion module in recent method UniTR~\cite{wang2023unitr}, we observe a degradation in performance on nuScenes (see Tab.~\ref{tab:med_linear}). Then what challenges prevent their effective use?

We believe that the core challenge lies in how to effectively arrange the feature elements from the two modalities in a unified space.
One intuitive approach is to use the BEV space as the unified coordinate system, a strategy that has been shown to be effective in previous Transformer-based local fusion models~\cite{liu2023bevfusion}. However, directly compressing the features into 3D space may reduce the efficacy of fusion, as it leads to the loss of height information that is crucial for global interaction. To address this issue, we propose using the image plane as the unified space, projecting the raw voxel features derived from the point clouds onto this space. Our study further reveals that the current voxel compression and merging techniques in LiDAR backbones introduce deviations in height information due to quantization errors in discrete space. This, in turn, leads to inevitable errors during multi-modality alignment (see Fig.~\ref{fig:method_proj}). To mitigate this, we propose a height-fidelity LiDAR encoding method, which performs voxel compression in continuous space, thereby retaining more accurate and reliable height information. With the constructed unified space and height-aware representation, we introduce a Hybrid Mamba Block (HMB) that simultaneously and efficiently learns mutual information at both local and global scales.

By incorporating the proposed method into a new framework, our model achieves state-of-the-art performances with NDS score of 75.0 on nuScenes~\cite{caesar2020nuscenes} validation dataset. Moreover, our framework demonstrates a $1.5\times$ faster speed compared to recent top method IS-FUSION~\cite{yin2024fusion}.

In summary, our main contributions are as follows: 1) We introduce the Hybrid Mamba Block to enable efficient dense global multi-modality fusion, which is compatible with different linear attention. 2) We propose a height-fidelity LiDAR encoding technique that retains more reliable height information, thereby improving the accuracy of camera-LiDAR alignment in a unified space. 3) For the first time, we demonstrate that a framework utilizing pure linear operations for fusion can achieve SOTA performance in camera-LiDAR multi-modal 3D object detection.

\begin{table}[!t]
\footnotesize
\centering
\begin{tabular}{lcccc}
\toprule

\textbf{Framework} & \textbf{Efficient} &\textbf{Far-sighted} & \textbf{Omniscient} \\

\midrule
Dense Local & \checkmark & \xmark & \xmark \\
Sparse Global & \checkmark & \checkmark & \xmark\\
Quasi Dense Global & \checkmark   & \xmark & \xmark \\
Dense Global \textbf{(Ours)} &  \checkmark & \checkmark & \checkmark \\
\bottomrule
\end{tabular}
\vspace{-0.1cm}
\caption{Analysis of different fusion strategies.}
\label{tab:categorization}
\vspace{-0.5cm}
\end{table}

\section{Related Work}

\begin{figure*}[t]
    \centering
    \footnotesize
    \includegraphics[width=0.98\linewidth]{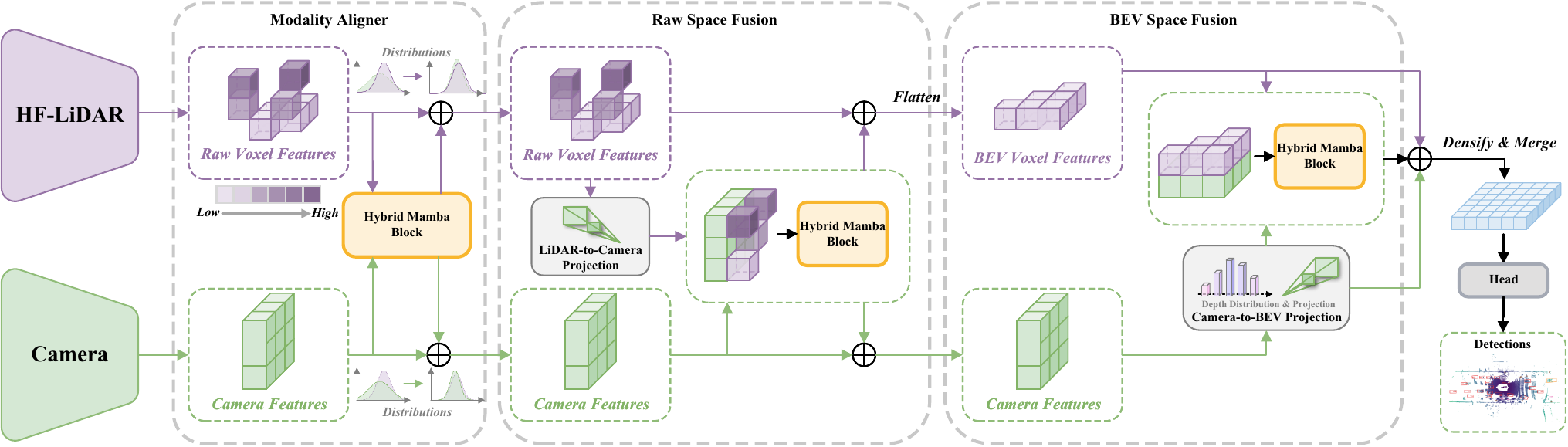}
    \vspace{-0.1cm}
    \caption{\textbf{Overview of our framework.} We propose a novel height-fidelity LiDAR encoding method, which preserves more accurate height information within voxel features, thereby enhancing the precision of multi-modal alignment. Then, we introduce the Hybrid Mamba Block (HMB), designed to: 1) align the latent-space distributions across modalities, 2) facilitate cross-modal information exchange between raw voxel and camera features, and 3) perform efficient multi-modal representation fusion and compression within BEV space. 
 }
    \label{fig:method_main}
    \vspace{-0.6cm}
\end{figure*}

\noindent\textbf{3D Object Detection with Cameras-LiDAR Fusion.} 

3D object detection using cameras and LiDAR is critical for autonomous vehicles. Early work focused on independent methods for each sensor type~\cite{wang2021fcos3d,wang2022detr3d,li2022bevformer,philion2020lift,huang2021bevdet,li2023bevdepth,han2024exploring,liu2022petr,liu2023sparsebev,lin2023sparse4d,qi2017pointnet++}. To improve driving safety, researchers explored fusion strategies that leverage both modalities’ complementary strengths. Early fusion~\cite{huang2020epnet,sindagi2019mvx,vora2020pointpainting,wang2021pointaugmenting,yin2021multimodal} augments raw point clouds with aligned image features. Middle fusion~\cite{liang2022bevfusion,liu2023bevfusion} extracts mutual information in a high-dimensional representation space~\cite{wang2023unitr,yin2024fusion,liu2023bevfusion}, enabling richer feature learning. Late fusion~\cite{zhou2023sparsefusion,zhang2024sparselif} aligns and merges features from predicted boxes. Revisiting these methods reveals that efficient long-range context learning remains a major bottleneck under strict latency constraints.

\noindent\textbf{Long Range Modeling.} Capturing long-range dependencies is essential for learning robust representations. In the context of language and vision foundation models, Transformers~\cite{vaswani2017attention} and large convolutional kernels~\cite{ ding2022scaling, rao2022hornet, liu2022more} have been proposed to address this challenge. However, the direct application of these operators to 3D detection framework presents significant difficulties. Specifically, conventional convolutional kernels struggle with the sparse nature of LiDAR data, while the global attention mechanism in Transformers incurs a prohibitively high quartic computational cost. To mitigate these challenges, window-based methods~\cite{wang2023unitr, yin2024fusion} have been introduced, which perform global fusion exclusively among window features, thereby reducing latency. However, this approach inevitably sacrifices some detail during long-range feature modeling.

\noindent\textbf{Linear Attention. }
To mitigate the quadratic complexity of vanilla attention, linear attention~\cite{katharopoulos2020transformers} was proposed and has shown strong gains~\cite{sun2023retentive,peng2023rwkv}. State space models (SSMs), a variant of linear attention~\cite{han2025demystify}, demonstrate superior performance. They have been applied in vision tasks~\cite{zhu2024vision,liu2024vmamba}, point cloud processing~\cite{liang2024pointmamba,zhang2024point,zhang2024voxel,liu2024lion}, and RGB-T fusion~\cite{gao2024mambast}. Yet fusing 2D camera and 3D LiDAR remains a significant challenge due to inherent disparities between these modalities.  DRAMA~\cite{yuan2024drama} uses Mamba to merge LiDAR and camera BEV features but compresses point clouds into BEV maps, sacrificing height information and relying on CNN backbones. In this work, we for the first time adapt pure linear attnetion to capture long and short-term dependencies in multi-modal 3D detection for autonomous driving, which enables flow of long-range information from single modality representation learning to multi-modal fusion.

\section{Method}

\subsection{Analysis of Different Linear Attention}
\label{sec:preliminaries}

In point cloud processing, non-uniform spatial intervals arising from occlusion and varying distances pose significant challenges for conventional linear attention methods. 
Some linear attention methods (\textit{e.g.}, Criss-Cross Attention~\cite{huang2019ccnet}) require continuous 2D inputs, making them unsuitable for sparse point clouds, while others (\textit{e.g.}, RWKV~\cite{peng2023rwkv}) rely on fixed space/time intervals. In contrast, Mamba leverages HiPPO~\cite{gu2020hippo} to capture long-range dependencies without relying on explicit space/time priors, which is an essential feature for handling variable spacing in sparse point clouds.
As illustrated in Tab.~\ref{tab:med_linear}, replacing the window-based transformer in UniTR~\cite{wang2023unitr} with linear attention results in a noticeable performance drop.
Although Mamba also experiences some decline, it maintains a smaller performance drop compared to other methods, demonstrating its relative robustness.
Based on the above analysis, these findings underscore Mamba’s advantages, motivating its adoption for our main experiments. The following sections detail our Mamba-based approach.

\begin{table}[t]
\centering
\setlength{\tabcolsep}{4pt}
\scalebox{0.8}{
\begin{tabular}{l|c |c| c| c}
\hline
Components&vanilla&improvements &+ours&improvements \\ 

\hline
RetNet~\cite{sun2023retentive}&64.6/68.3
&\textcolor[RGB]{61,145,64}{-6.0}/\textcolor[RGB]{61,145,64}{-4.6}
&70.7/73.1
&\textcolor{red}{+0.1}/\textcolor{red}{+0.2}\\
\hline
RWKV~\cite{peng2023rwkv}&64.9/68.8
&\textcolor[RGB]{61,145,64}{-5.7}/\textcolor[RGB]{61,145,64}{-4.1}
&70.8/73.3
&\textcolor{red}{+0.2}/\textcolor{red}{+0.4}\\ 
\hline
Mamba~\cite{gu2023mamba}&65.8/70.4&\textcolor[RGB]{61,145,64}{-4.8}/\textcolor[RGB]{61,145,64}{-2.5}&71.9/74.3&\textcolor{red}{+1.3}/\textcolor{red}{+1.4}\\
\hline
\end{tabular}

}
\vspace{-0.1cm}
\caption{Comparison of different linear attention methods on 50\% nuScenes \textit{v.s.} UniTR~\cite{wang2023unitr} with 70.6/72.9 (mAP/NDS).}
\label{tab:med_linear}
\vspace{-0.6cm}
\end{table}

\subsection{Overall Architecture}
\label{sec:overall}
\begin{figure}[t]
    \centering
    \footnotesize
    \includegraphics[width=0.98\linewidth]{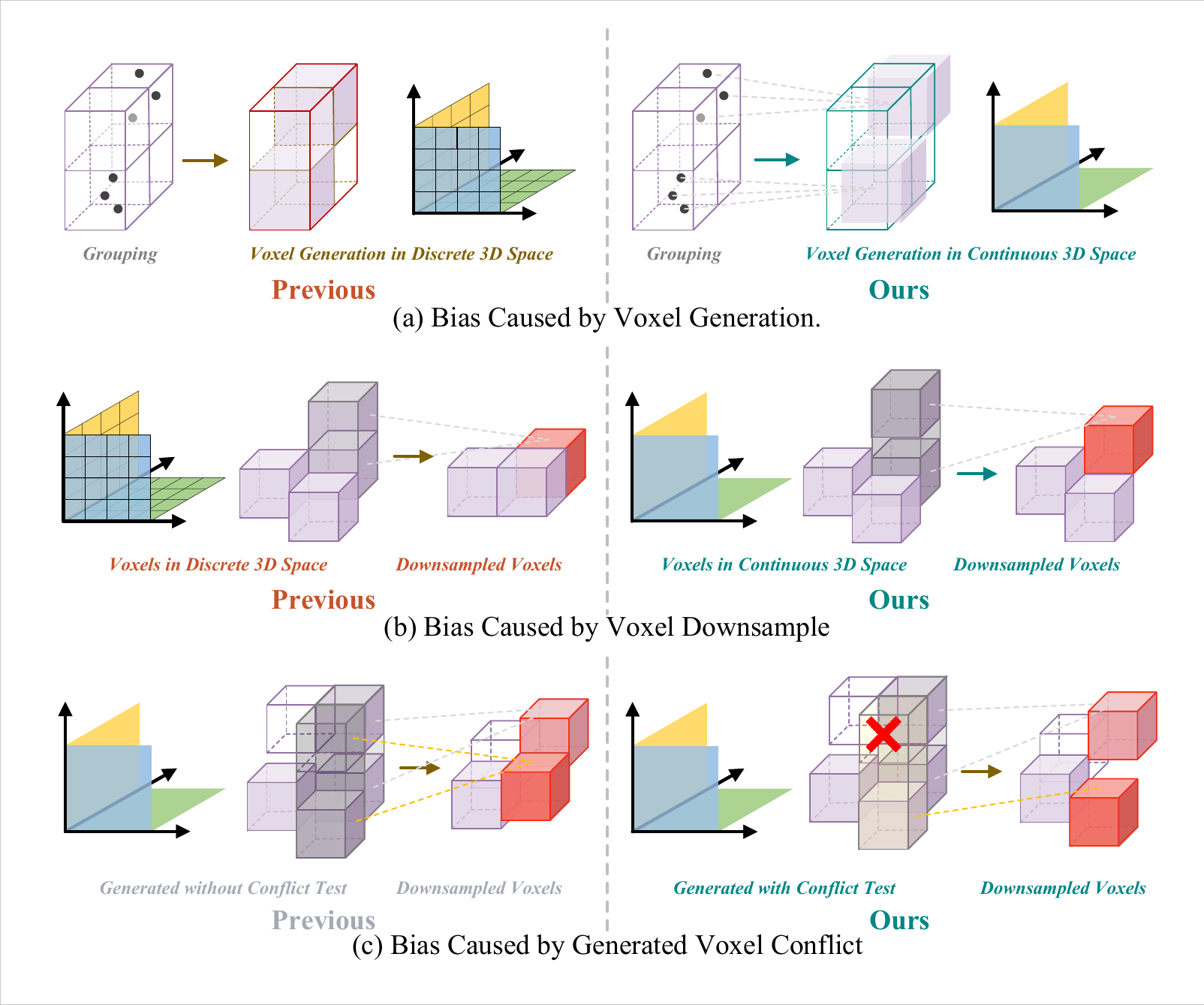}
    \vspace{-0.1cm}
    \caption{\textbf{Height-Fidelity LiDAR Encoding.} (a-b) Height information loss caused by quantization error in discrete space. (c) Height deviation introduced by the generated voxels in LION~\cite{liu2024lion}.}
    \label{fig:method_height}
    \vspace{-0.5cm}
\end{figure}

Fig.~\ref{fig:method_main} illustrates the architecture of our proposed framework, designed for dense global fusion in multimodal 3D detection. Firstly, we extract visual and sparse point cloud features from multi-view images and point clouds. In this work, we propose the height-fidelity encoding strategy to preserve more detailed height information of point cloud features. This strategy is specifically designed to enhance the quality of multi-modal feature alignment during the fusion of LiDAR and camera features, as illustrated in Fig.~\ref{fig:method_height} and Sec.~\ref{sec:height}. Secondly, the modality aligner harmonizes the feature distributions of the two modalities, stabilizing the multi-modal fusion process. Thirdly, to learn the mutual information between the modalities, we project their features into unified coordinate systems in two ways: 1) height-fidelity LiDAR features are projected into the image space and reordered with the raw visual feature embeddings; 2) visual features are projected into the BEV space and reordered with the point cloud features. These reordered features are fused using our proposed Hybrid Mamba Block (see Sec.~\ref{sec:hybrid_block}). Finally, the detection head predicts the results using the fused features from the BEV space.

\subsection{Hybrid Mamba Block}
\label{sec:hybrid_block}

To achieve dense global fusion, we introduce the Hybrid Mamba Block, which combines local and global Mamba modules to integrate information across all modalities in the entire scenario. 
The local module captures fine-grained features, while the global module enhances the understanding of the broader driving environment. Notably, the proposed Hybrid Mamba Block can either independently process the features from each modality or jointly fuse their information within a unified coordinate system, offering flexibility to adapt to various fusion requirements.

\begin{figure}[t]
    \centering
    \footnotesize
    \includegraphics[width=0.98\linewidth]{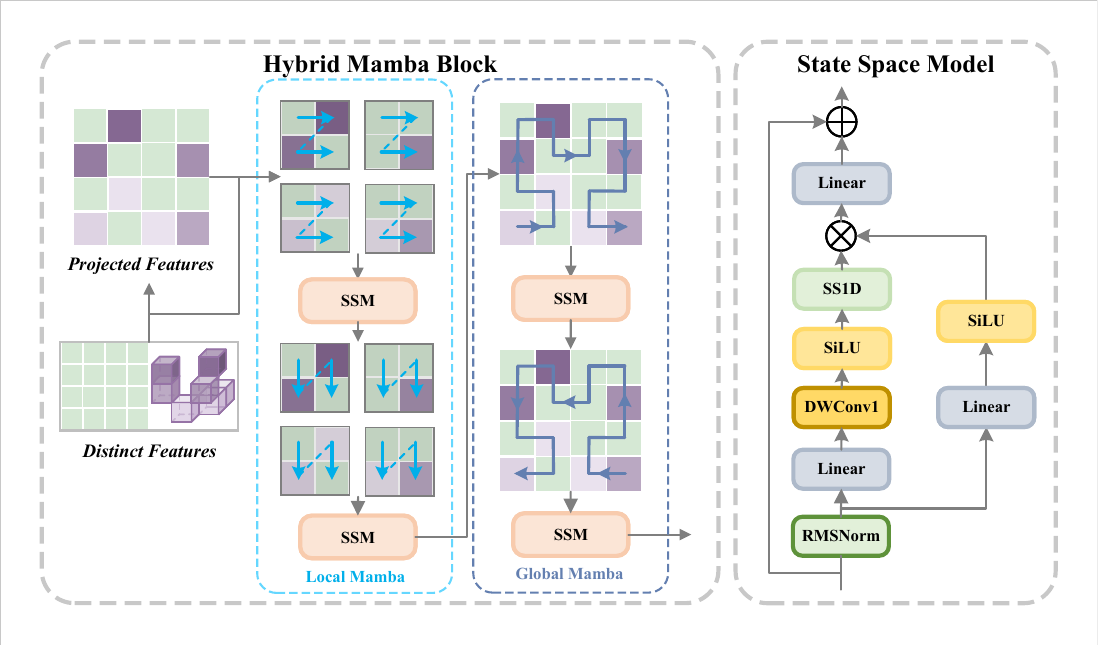}
    \vspace{-0.1cm}
    \caption{\textbf{Illustration of the Hybrid Mamba Block.}  The block first employs the \textbf{\textcolor[RGB]{0, 175, 234}{Local Mamba}} to enable localized feature interaction around each spatial position. Subsequently, the \textbf{\textcolor[RGB]{99, 127, 176}{Global Mamba}} facilitates interaction across the entire space.}

    \label{fig:method_mamba}
    \vspace{-0.5cm}
\end{figure}

\noindent\textbf{Preliminaries.} Rooted in linear systems theory~\cite{hespanha2018linear}, the State Space Model (SSM)~\cite{gu2021combining,gu2023mamba,gu2021efficiently,dao2024transformers} provides a robust framework for representing dynamical systems. 
The continuous-time SSM is:
\begin{align} 
\mathbf{h}'(t) &= \mathbf{A} \mathbf{h}(t) + \mathbf{B} \mathrm{x}(t), \label{eq:ssm1} \\
\mathbf{y}(t) &= \mathbf{C}^\top \mathbf{h}(t) + \mathbf{D} \mathbf{x}(t), \label{eq:ssm2}
\end{align}
where $\mathbf{h}(t) \in \mathbb{R}^{ C}$ is the hidden state, $\mathbf{y}(t) \in \mathbb{R}^L$ is the output, $\mathbf{A}$ governs system dynamics, $\mathbf{B}$ insert the input $\mathbf{x}(t)$ to the state, $\mathbf{C}$ projects the state to output, and $\mathbf{D}$ allows residual connection. 
$\mathbf{A}$ based on HiPPO~\cite{gu2020hippo}  in Mamba effectively captures long-range dependencies by transforming global features into a compressed representation.

\noindent\textbf{Global Mamba.} It is well-established that Mamba was originally designed to process 1D sequential data. To extend its applicability to 2D camera/BEV space and raw 3D space, we adopt the Hilbert curve~\cite{hilbert1935stetige}, as used in prior work~\cite{li2024mamba24}, to serialize the feature tokens from both modalities. Specifically, the Hilbert curve traverses the feature embeddings of both the camera and LiDAR modalities, serving as a space-filling curve renowned for preserving spatial locality, as illustrated in Fig.~\ref{fig:method_mamba}. This approach ensures efficient traversal and seamless integration of heterogeneous information in the entire 3D space.

More precisely, given the embeddings from single- or multi-modalities feature $\mathbf{F}$ with their corresponding coordinates $\mathcal{C}=\{(x_{\mathrm{i}}, y_{\mathrm{i}}, z_{\mathrm{i}})\}_{\mathrm{i}=1}^{N_{\rm feat}}$
, we compute the Hilbert indices $\mathcal{P}_{\mathrm{h}}=\{h_{\mathrm{i}}\}_{\mathrm{i}=1}^{N_{\rm feat}}$
for each embedding following ~\cite{skilling2004programming},
\begin{equation}
    h_{\mathrm{i}} = \mathrm{HilbertIndex}(x_{\mathrm{i}}, y_{\mathrm{i}}, z_{\mathrm{i}}),
\end{equation}
where $\mathrm{HilbertIndex}(\cdot)$ maps the 3D coordinate to a 1D Hilbert curve index, so that we can reorder the embeddings in features $\mathbf{F}$. 
Besides, we use positional embedding to preserve information about the original coordinates,
\begin{equation}
    \mathbf{F}' \leftarrow \mathbf{F} + \mathrm{PosEmbedding}(\mathcal{C}).
\end{equation}
Since the receptive field of Mamba is unilateral, we process the reordered embeddings in a bi-directional way,
\begin{equation}
    \mathbf{F}_{\mathrm{global}} = \mathrm{Mamba}(\mathbf{F}',\mathcal{P}_{\mathrm{h}} ).
\end{equation}
In this manner, each token comprehensively integrates information from all available features, effectively capturing long-range dependencies across the entire space.

\begin{figure*}[t]
    \centering
    \footnotesize
    \includegraphics[width=0.98\linewidth]{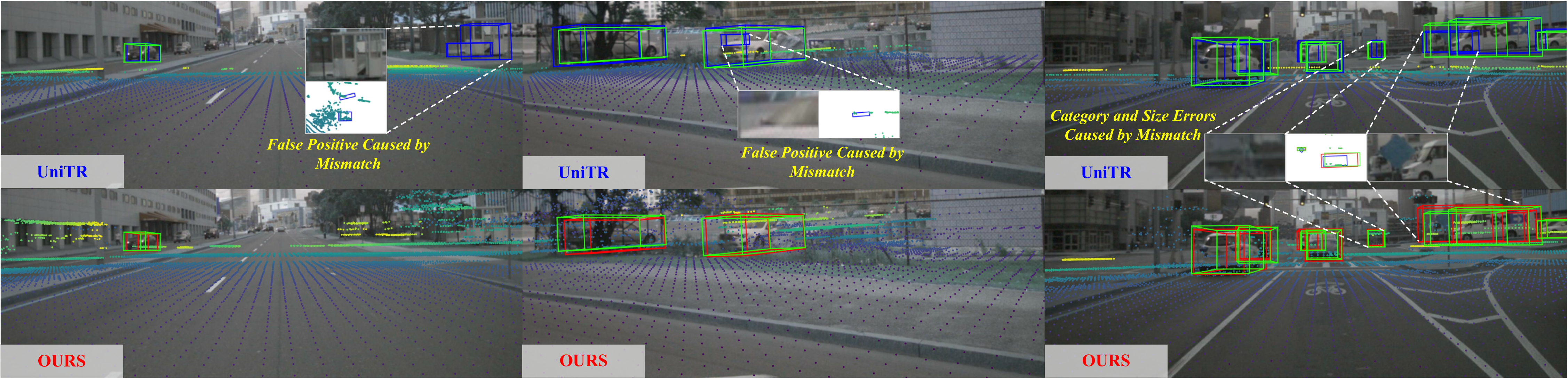}
    \vspace{-0.1cm}
    \caption{\textbf{Voxel projection and detection results on images.} Comparison of our projection method (bottom), enhanced by our Height-Fidelity LiDAR encoding, with previous methods (top). The figure demonstrates that projection errors introduce false positives, category errors, and unsatisfactory scales. The \textcolor[RGB]{33, 255, 5}{\textbf{green}}, \textcolor[RGB]{0,0,255}{\textbf{blue}} and \textcolor[RGB]{255,0,0}{\textbf{red}}
    boxes indicate ground truth, UniTR's output, and our results, respectively.}
    \label{fig:method_proj}
    \vspace{-0.5cm}
\end{figure*}

\noindent\textbf{Local Mamba.} Representation learning in local regions has been extensively studied, which is demonstrated to be a valuable complement to global features. Therefore, we also adopt Mamba in small regions to capture local structure. As depicted in Fig.~\ref{fig:method_mamba}, we segment the input features into non-overlapping regions of size $w \times w$. For each feature embedding $\mathbf{f}_{\mathrm{i}} \in \mathbb{R}^{C}$ located at coordinates $(x_{\mathrm{i}}, y_{\mathrm{i}}, z_{\mathrm{i}})$, we compute its region index $r_{\mathrm{i}}$ and its positions $(x'_{\mathrm{i}}, y'_{\mathrm{i}})$ in the local region as follows:
\begin{equation}
\begin{cases}
\begin{aligned}
    r_{\mathrm{i}} &= \left\lfloor \frac{x_{\mathrm{i}}}{w} \right\rfloor \times \left\lceil \frac{y_{\mathrm{i}}}{w} \right\rceil + \left\lfloor \frac{y_{\mathrm{i}}}{w} \right\rfloor, \\
    (x'_{\mathrm{i}}, y'_{\mathrm{i}}) &= (x_{\mathrm{i}} \bmod w,\, y_{\mathrm{i}} \bmod w).
\end{aligned}
\end{cases}
\end{equation}
By combining the region index and the in-region positions, we obtain the new positions $\mathcal{P}_{\mathrm{r}} = \{ r_{\mathrm{i}}, x'_{\mathrm{i}}, y'_{\mathrm{i}} \}_{\rm i=1}^{N_{\rm feat}}$
for each feature embedding. We then apply the Mamba to aggregate information within each region using these new positions, yielding locally enriched features $\mathbf{F}_{\mathrm{local}}$ as follows:
\begin{equation}
    \mathbf{F}_{\mathrm{local}} = \mathrm{LocalMamba}(\mathbf{F},\, \mathcal{P}_{\mathrm{r}}).
\end{equation}
To facilitate comprehensive information exchange, we perform fusion along both the $x$ and $y$ directions (see Fig.~\ref{fig:method_mamba}).

\subsection{Height-Fidelity LiDAR Encoding}
\label{sec:height}

As shown in Fig.~\ref{fig:method_height}(a), during voxel generation in the LiDAR backbone, each voxel’s position is represented solely by the 3D coordinates of its centroid, disregarding the internal point distribution within the voxel. Conventional voxel feature integration repeatedly downscales features along the height dimension, which determines the 3D position of new voxels by averaging the coordinates of the merged ones. This process results in further loss of height information, as illustrated in Fig.~\ref{fig:method_height}(b). Fig.~\ref{fig:method_proj} shows the negative effects of height information loss during multi-modality alignment, which eventually degrade detection accuracy and quality.

To mitigate the height information loss caused by quantization errors in discrete space, we calculate the voxel coordinates directly in continuous 3D space. 
\begin{equation}
    \mathcal{C}_{\mathrm{P}_{0}} = \mathrm{ScatterMean}(\mathbf{c}, \boldsymbol{v}_{0}),
\end{equation}
where $\boldsymbol{v}_{0}$ represents the voxel assigned for each point, 
$\mathbf{c} \in \mathbb{R}^{N_{\rm point} \times 3}$
is the coordinates of point clouds. Then, we also perform the downsampling operation using raw coordinates in continuous 3D space. Specifically, as depicted in Fig.~\ref{fig:method_height}(b), during voxel merging at stage $s$, we compute the coordinates $\mathcal{C}_{\mathrm{P}_{s}}$ of the corresponding voxels in raw space based on the contained point clouds,
\begin{equation}
    \mathcal{C}_{\mathrm{P}_{s}} = \mathrm{ScatterMean}(\mathcal{C}_{\mathrm{P}_{s-1}}, \boldsymbol{v}_{s-1}).
\end{equation}

In addition to biases introduced by discretized coordinates, voxel generation also introduces deviation in our baseline LiDAR backbone LION~\cite{liu2024lion}. Specifically, it selects the top $k$ salient features with coordinates $\{x_{\rm i}, y_{\rm i}, z_{\rm i}\}_{\rm i=1}^{k}$, and generate new voxels at positions $\{(x_{\rm i} \pm 1, y_{\rm i} \pm 1, z_{\rm i})\}_{\rm i=1}^{k}$. However, as illustrated in Fig.~\ref{fig:method_height}(c), this approach may introduce errors during the downsampling process due to overlapping or conflicting voxels. To address this issue, we exclude voxels that are likely to merge with surrounding voxels, ensuring that only distinct, non-overlapping voxels are retained,
\begin{equation}
    \left\{ \left( x_{\rm i} \pm 1,\, y_{\rm i} \pm 1,\, \left\lfloor \frac{z_{\rm i}}{m} \right\rfloor \right) \right\}_{\rm i=1}^{k} \notin \mathcal{C}_{\mathrm{P}_{s}},
\end{equation}
where $m$ is the downsampling scale.

\begin{table*}[!t]
\centering

\scalebox{0.9}{
\begin{tabular}{l|c|c|c|c|c|c|c|c}
\toprule

\textbf{Methods} & \textbf{Present at} & \textbf{Resolution}&\textbf{Modality} & \textbf{NDS (val)} & \textbf{mAP (val)} & \textbf{NDS (test)} & \textbf{mAP (test)} & \textbf{FPS} \\
\hline
\hline

PointPainting \cite{vora2020pointpainting} & CVPR'20 &-& C+L & 69.6 & 65.8 & - & - & -\\
PointAugmenting \cite{wang2021pointaugmenting} & CVPR'21 &$1600\times 900$& C+L & - & - & 71.0 & 66.8 &  -\\
MVP \cite{yin2021multimodal} & NeurIPS'21 &$1600\times 900$& C+L & 70.0 & 66.1 & 70.5 & 66.4 & - \\

TransFusion \cite{bai2022transfusion} & CVPR'22 &$800\times 448$& C+L & 71.3 & 67.5 & 71.6 & 68.9 & - \\
AutoAlignV2 \cite{chen2022deformable} & ECCV'22 &$1440\times 800$& C+L & 71.2 & 67.1 & 72.4 & 68.4 &  -\\
UVTR \cite{li2022unifying} & NeurIPS'22 &$1600\times 900$& C+L & 70.2 & 65.4 & 71.1 & 67.1 & - \\
BEVFusion (PKU)~\cite{liang2022bevfusion} & NeurIPS'22 &$704\times 256$& C+L & 71.0 & 67.9 & 71.8 & 69.2 &0.8 \\
DeepInteraction~\cite{yang2022deepinteraction} & NeurIPS'22 &$800\times 448$& C+L & 72.6 & 69.9 & 73.4 & 70.8 &2.6\\
BEVFusion (MIT)~\cite{liu2023bevfusion} & ICRA'23 &$704\times 256$& C+L & 71.4 & 68.5 & 72.9 & 70.2 &4.2
\\
CMT~\cite{yan2023cross} & ICCV'23 &$1600\times 640$& C+L & 72.9 & 70.3 & 74.1 & 72.0 &2.8\\
SparseFusion~\cite{zhou2023sparsefusion}  & ICCV'23 &$704\times 256$& C+L & 72.8 & 70.4 & 73.8 & 72.0 &5.3 \\
ObjectFusion~\cite{cai2023objectfusion}  & ICCV'23 &$704\times 256$& C+L & 72.3 & 69.8 & 73.3 & 71.0 &- \\
UniTR~\cite{wang2023unitr}  & ICCV'23 &$704\times 256$& C+L & 73.3 & 70.5 & 74.5 & 70.9 &4.9\\

\hline
IS-FUSION~\cite{yin2024fusion}  & CVPR'24 &$1056\times 384$& C+L & 74.0 &   \textcolor[RGB]{255,0,0}{\textbf{72.8}} & \textcolor[RGB]{0,0,255}{\textbf{75.2}} & \textcolor[RGB]{0,0,255}{\textbf{73.0}} &3.2\\
DAL~\cite{huang2025detecting} & ECCV'24 & $1056\times 384$ & C+L & 74.0 & 71.5 & 74.8 &72.0 & 4.3\\
SparseLIF~\cite{zhang2024sparselif} & ECCV'24 & $1600\times 640$ & C+L &\textcolor[RGB]{0,0,255}{\textbf{74.6}}   & 71.2&-&-&2.9\\
\hline

\cellcolor{lightgray!15}MambaFusion-Lite &\cellcolor{lightgray!15}Ours &\cellcolor{lightgray!15}$704\times 256$&\cellcolor{lightgray!15}C+L &\cellcolor{lightgray!15}74.0&\cellcolor{lightgray!15}71.6&\cellcolor{lightgray!15}75.0 &\cellcolor{lightgray!15}72.0&\cellcolor{lightgray!15}5.4 \\
\cellcolor{lightgray!40}MambaFusion-Base &\cellcolor{lightgray!40}Ours &\cellcolor{lightgray!40}$704\times 256$&\cellcolor{lightgray!40}C+L &  \cellcolor{lightgray!40}\textcolor[RGB]{255,0,0}{\textbf{75.0}} &\cellcolor{lightgray!40}\textcolor[RGB]{0,0,255}{\textbf{72.7}} &\cellcolor{lightgray!40}\textcolor[RGB]{255,0,0}{\textbf{75.9}} &  \cellcolor{lightgray!40}\textcolor[RGB]{255,0,0}{\textbf{73.2}}&\cellcolor{lightgray!40}4.7\\
\specialrule{.1em}{0em}{0em}
\end{tabular}
}
\vspace{-0.1cm}
\caption{Comparison on the nuScenes validation and test datasets~\cite{caesar2020nuscenes}. The modalities used are Camera (C) and LiDAR (L). The highest scores are in \textcolor[RGB]{255,0,0}{\textbf{red}}, while the second-highest are in \textcolor[RGB]{0,0,255}{\textbf{blue}}. Note that SparseLIF does not provide test set results for single-frame models.}
\label{tab:valtest}
\vspace{-0.5cm}
\end{table*}

\subsection{Multi-modality Representation Learning}
\label{sec:fusion}

This section provides a detailed description of the development of a fusion framework incorporating the proposed modules. First, we align the feature distributions of point cloud and image modalities by passing them through a shared HMB.
This process is akin to the behavior of "normalization" layers. After achieving consistent feature distributions, we perform fusion in both the raw 3D space and the BEV space. In the raw space, we project the 3D voxel features onto the image planes, facilitating the fusion of the image and point cloud features. In the BEV space, we use the Lift-Splat-Shoot (LSS) transformation to project image features into the BEV space and fuse them with the point cloud features, creating a unified multi-modal representation. This fusion process is carried out efficiently by downsampling BEV features and preserving only relevant information in the point cloud features, ultimately providing a rich representation for the detection task.

\section{Experiments}
\subsection{Experimental Setup}

\noindent\textbf{Datasets and Evaluation Metrics.}
We evaluate our framework on the nuScenes~\cite{caesar2020nuscenes} dataset, which comprises 40,157 annotated samples collected from six cameras providing a full 360-degree field of view and a 32-beam LiDAR sensor. For evaluation, we employ Mean Average Precision (mAP) and the nuScenes Detection Score (NDS).

\noindent\textbf{Implementation Details.}
Following UniTR~\cite{wang2023unitr}, we use a voxel resolution of $[0.3, 0.3, 0.25]$. The spatial range is $[-54.0m, 54.0m]$ for $x$ and $y$ axes, and $[-5.0m, 3.0m]$ for $z$ axis. Multi-view images are used at a resolution of $704 \times 256$. The LiDAR and camera tokenizers are partly adapted from VMamba~\cite{liu2024vmamba} and LION~\cite{liu2024lion}. Adam optimizer and one-cycle learning rate policy~\cite{smith2017cyclical} are used during training. Our model is trained for 10 epochs with a maximum learning rate of $1.5 \times 10^{-3}$ on four A100 GPUs.

\begin{table}[t]
\centering
\setlength{\tabcolsep}{3pt}
\scalebox{0.85}{
\begin{tabular}{c| c| c c c | c | c c}
\toprule
 &  Baseline-LC  & HMB-M & HMB-R &HMB-B & HFL & mAP & NDS \\
\hline
\hline

\ding{172}& \checkmark & &&&   &67.7& 70.9\\

\ding{173}& \checkmark & \checkmark & &  &&68.6&71.6 \\
\ding{174}& \checkmark & & \checkmark &  & &66.9& 70.2\\
\ding{175}& \checkmark & &  & \checkmark & &69.6& 72.0\\
\ding{176}& \checkmark & & \checkmark &  &\checkmark &70.3& 72.7\\

\ding{177}& \checkmark & \checkmark & &\checkmark& & 70.6&73.1 \\
\ding{178}& \checkmark &\checkmark & \checkmark &  &\checkmark &71.4& 73.6\\
\ding{179}& \checkmark && \checkmark & \checkmark &\checkmark &71.2& 73.7\\
\ding{180}& \checkmark & \checkmark & \checkmark& \checkmark&\checkmark  & 71.9& 74.3\\
\bottomrule
\end{tabular}}
\vspace{-0.1cm}
\caption{Ablation study of each module. 'Baseline-LC' refers to a simplified version of the model, 'HFL' denotes the height-fidelity LiDAR encoding, and 'HMB-X' represents the Hybrid Mamba Block applied at different stages, as described in Sec.~\ref{sec:fusion}.}
\label{tab:ablmain}
\vspace{-0.5cm}
\end{table}

\subsection{Comparison with State-of-the-Art Methods}
As shown in Tab.~\ref{tab:valtest}, we present results on the nuScenes validation and test sets, obtained without test-time augmentations or model ensembling. Our Lite model outperforms the previous quasi-dense global fusion method UniTR~\cite{wang2023unitr} by 1.1 mAP and 0.7 NDS with 10\% faster inference speed. Compared with the recent state of the arts, our Base model achieves 75.0 NDS on the validation dataset, surpassing all compared methods. 
Moreover, compared to the most recent IS-Fusion~\cite{yin2024fusion} and SparseLIF~\cite{zhang2024sparselif}, our method achieves a 50\% and 62\% increase in inference speed.
Overall, our method strikes an impressive balance between performance and efficiency.
Detailed per-category and other criteria like mASE are provided in the supplementary material.

\begin{table*}[t!]
\centering
\setlength{\tabcolsep}{4pt} 

\begin{tabular}{ccc}

\begin{minipage}{0.32\textwidth}
\centering
\scalebox{0.85}{
\begin{tabular}{lcc}
\toprule
{Component} & {mAP} & {NDS} \\
\hline
\hline
Local only & 70.5& 72.1 \\
Global only &71.5  & 73.7\\
Local and Global &71.9  &74.3 \\
\bottomrule
\end{tabular}
}
\subcaption{Local and global fusion.}
\label{tab:hybrid_mamba_local_global}
\vspace{-0.25cm}
\end{minipage}
&
\begin{minipage}{0.32\textwidth}
\centering

\scalebox{0.85}{
\begin{tabular}{lcc}
\toprule
{Sorting Paradigm} & {mAP} &{NDS} \\
\hline
\hline
Coordinate Order &  71.2&  73.5\\
Z-Order Curve~\cite{morton1966computer} & 71.5& 74.0\\
Hilbert Curve~\cite{hilbert1935stetige} &71.9&74.3 \\
\bottomrule
\end{tabular}
}
\subcaption{Different sorting paradigm in global Mamba.}
\label{tab:hybrid_mamba_space_filling}
\vspace{-0.25cm}
\end{minipage}
&

\begin{minipage}{0.32\textwidth}
\centering
\scalebox{0.85}{
\begin{tabular}{lcc}
\toprule
{Component} & {mAP} & {NDS} \\
\hline
\hline
Baseline &  69.5&  72.2\\
+ xy in local & 70.9& 73.0\\
+ bidirectional in global  & 71.9  & 74.3\\
\bottomrule
\end{tabular}
}
\subcaption{Ablation on serialization combination.}
\label{tab:hybrid_mamba_aggregation}
\vspace{-0.3cm}
\end{minipage}

\end{tabular}
\caption{Design choices of hybrid Mamba block. We explore the impact of each detailed components.}
\label{tab:hybrid_mamba}
\vspace{-0.5cm}
\end{table*}

\subsection{Ablation Studies}
This section analyzes the impact of each proposed module. To expedite validation during ablation experiments, we use only half of the dataset, constructed by selecting every alternate frame. Evaluation is performed on the nuScenes validation set unless otherwise specified.

\subsubsection{Component-wise Ablation}
As presented in Tab.~\ref{tab:ablmain}, we conducted an ablation study to evaluate each component's contribution. The Baseline-LC model, integrating point cloud and image features following BEVFusion~\cite{liang2022bevfusion}, achieved an mAP of 67.7 and an NDS of 70.9 (\ding{172}). Incorporating  Hybrid Mamba Block (HMB) at various stages generally enhanced performance (\ding{173}\ding{175}). However, when fusing multi-modal features in the raw 3D space (\ding{174}), the performance degrades due to projection misalignments between modalities (Fig.~\ref{fig:method_proj}). Since Mamba is sequential, alignment errors would be amplified compared to Transformers. Introducing the height-fidelity LiDAR encoding (HFL) solves this issue (\ding{176} \textit{vs.}\ding{174}). Finally,  integrating HMB at different stages along with HFL further enhanced performance, achieving 71.9 mAP and 74.3 NDS.

\subsubsection{Analysis of Hybrid Mamba Block}
In this section, we study the effect of detailed design in the proposed Hybrid Mamba Block. As shown in Tab.~\ref{tab:hybrid_mamba}\subref{tab:hybrid_mamba_local_global}, we find that only using local Mamba leads to suboptimal results, while only employing the global Mamba yields significantly better performance with 1.0 gains in mAP and 1.6 gains in NDS. Combining local and global Mamba components further enhances the results, demonstrating their complementary benefits. Tab.~\ref{tab:hybrid_mamba}\subref{tab:hybrid_mamba_space_filling} presents the impact of different 3D space-filling curves, where the Hilbert Curve outperforms others in capturing spatial relationships. 
Lastly, as shown in Tab.~\ref{tab:hybrid_mamba}\subref{tab:hybrid_mamba_aggregation}, we investigate different serialization combinations within the Hybrid Mamba Block, where the bidirectional strategy demonstrates better performances.

\begin{table}[t!]
\centering

\scalebox{0.85}{
\begin{tabular}{c|c c c | c c}
\toprule
 &  Backbone  & VoxelVFE&Conflict Test & mAP & NDS \\
\hline
\hline
\ding{172} &  & & &  70.3& 73.1\\
\ding{173} & \checkmark & & &  71.2& 73.8\\

\ding{174} & \checkmark & \checkmark&  & 71.5 & 73.9\\
\ding{175} & \checkmark & \checkmark & \checkmark  & 71.9& 74.3 \\
\bottomrule
\end{tabular}}
\vspace{-0.1cm}
\caption{``Backbone'' denotes the phase of voxel integration (Fig.~\ref{fig:method_height}(a)). ``VoxelVFE'' refers to the stage when voxels are generated (Fig.~\ref{fig:method_height}(b)). ``Conflict Test'' refers to that in Fig.~\ref{fig:method_height}(c).}
\label{tab:exp_height}
\vspace{-0.5cm}
\end{table}

\subsubsection{Impact of Height-Fidelity LiDAR Encoding}
As shown in Fig.~\ref{fig:method_proj}, previous methods that omit height information fail to effectively project corresponding point clouds to regions where objects are presented, resulting in false positives, misclassifications, and erroneous estimations of object scale.
Quantitative analysis in Tab.~\ref{tab:exp_height} demonstrates that incorporating height information at different stages enhances performance. Specifically, adding height-aware encoding in the backbone improves the mAP by 0.9 and NDS by 0.7 compared to the baseline (\ding{173}$vs.$\ding{172}). \ding{174} and \ding{175} further improve performance by preserving height information during voxel generation ((Fig.~\ref{fig:method_height}(b))) and applying the conflict test strategy (Fig.~\ref{fig:method_height}(c)), where \ding{175} obtains an mAP of 71.9 and NDS of 74.3.
These findings confirm the effectiveness of our height-fidelity LiDAR encoding in improving feature alignment between LiDAR and image data, 
thereby enhancing overall detection performance.

\begin{table}[t!]
\centering
\setlength{\tabcolsep}{2pt}
\scalebox{0.85}{
\begin{tabular}{c|c c c | c c|c}
\toprule
 &   Component & Alignment  &Space& mAP & NDS&FPS\\
\hline
\hline
\ding{172} & Win-Transformer  & Discrete & Image+Frustum&68.9 & 71.9 &4.9\\
\ding{173} & Win-Transformer  & Discrete & Image+BEV& 69.5&  72.2& 3.2\\
\ding{174} & Win-Transformer  &  Continuous & Image+BEV&70.6&  72.9&3.1\\
\hline
\ding{175} & Vanilla Mamba & Discrete& Image+Frustum &65.8  & 70.4& 6.9\\
\ding{176} & Vanilla Mamba  &Discrete &Image+BEV & 68.0 & 70.7 &6.4\\
\ding{177} & Hybrid Mamba &Discrete &Image+BEV&  70.3& 73.1&4.8\\

\ding{178} & Hybrid Mamba  & Continuous &Image+BEV & 71.9 &74.3&4.7  \\
\bottomrule
\end{tabular}}
\vspace{-0.1cm}

\caption{Comparison with window-based Transformer. ``Vanilla Mamba'' denotes directly replacing the Transformer with Mamba. ``Discrete'' means using coordinates in discrete space, while ``Continuous'' refers to the continuous space employed in the height-fidelity LiDAR encoding. ``Image+Frustum'' corresponds to the coordinate space utilized in UniTR~\cite{wang2023unitr}, whereas ``Image+BEV'' pertains to the coordinate space described in Sec.~\ref{sec:fusion}.}
\label{tab:exp_transformer}
\vspace{-0.25cm}
\end{table}

\subsubsection{Comparison with Transformer}
In this section, we analyze the performance and efficiency of the window-based Transformer compared to our proposed model. As shown in Tab.~\ref{tab:exp_transformer}~\ding{172}, we replicate the method from UniTR~\cite{wang2023unitr}. Initially, we replace all Transformer layers with Mamba. While this substitution accelerates computation, it results in a significant decrease in performance (from 68.9 to 65.8 mAP in \ding{176}).
To address this issue, we change the fusion space to Image+BEV (\ding{176}) and incorporate the Hybrid Mamba Block along with the height-fidelity LiDAR encoding (\ding{177} and \ding{178}). These modifications lead to a substantial improvement in model performance.
Finally, we apply the approaches from \ding{176} and \ding{178} to the window-based Transformer to evaluate the generalizability of our proposed method. However, the window-based Transformer incurs higher computational costs and obtains lower performance due to its limited global perception capability (\ding{173} and \ding{174}).
Moreover, our proposed height-fidelity LiDAR encoding significantly enhances performance with minimal additional computational overhead (\ding{178} $vs.$ \ding{177}).

\begin{figure}[t]
    \centering
    \footnotesize
    \includegraphics[width=0.98\linewidth]{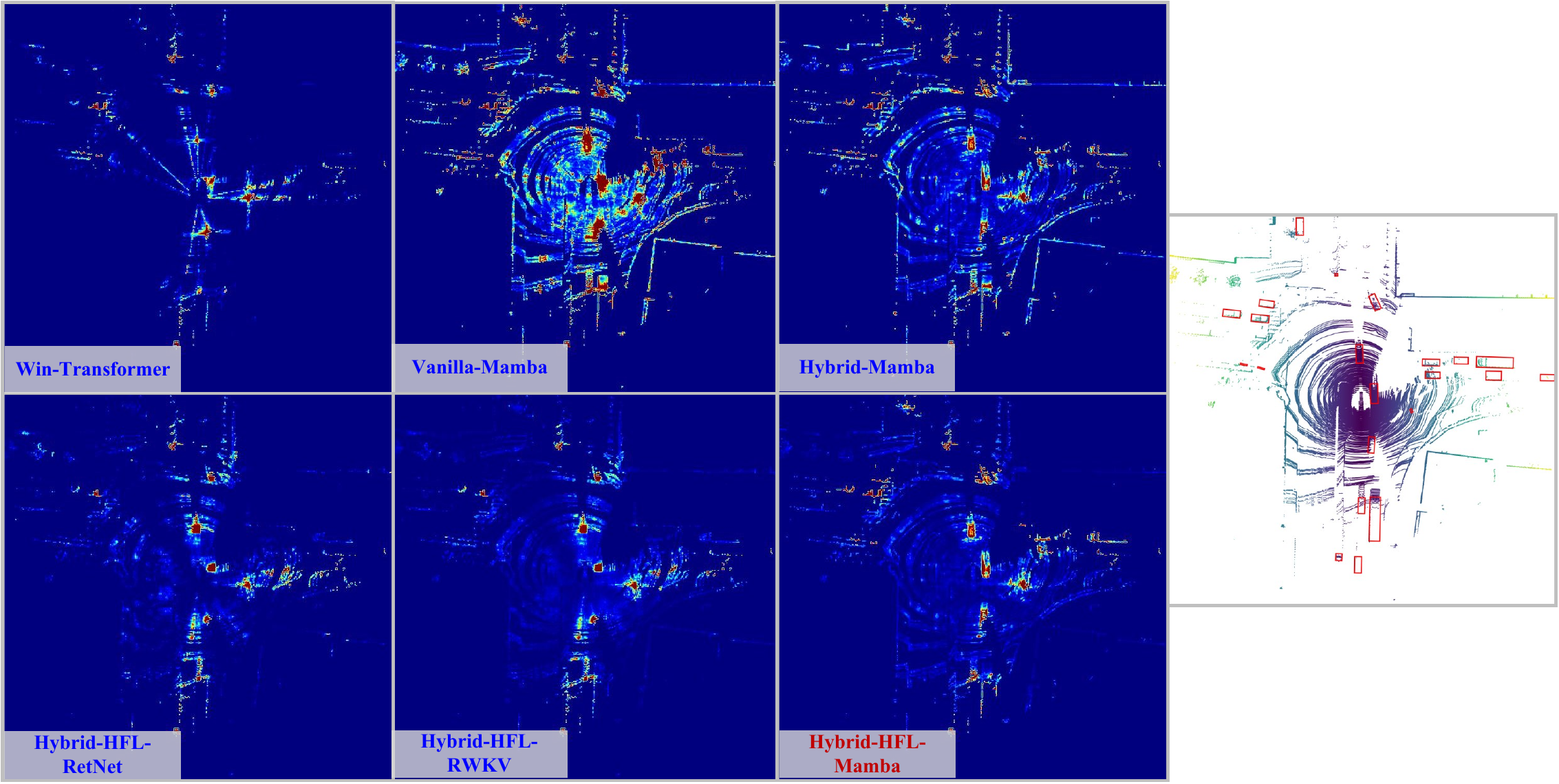}
    \vspace{-0.1cm}
    \caption{\textbf{The effective receptive fields (ERFs).}  ``Win'' denotes window partition, ``Hybrid'' means Hybrid Mamba (RetNet, RWKV) Block, ``HFL'' refers to Height-Fidelity LiDAR Encoding.}


    \label{fig:exp_heatmap}
    \vspace{-0.5cm}
\end{figure}

\subsubsection{ERFs and Analysis of Long-rang Modeling}

As shown in Fig.~\ref{fig:exp_heatmap}, we visualize the effective receptive fields (ERFs) to investigate how each approach captures long-range dependencies. Specifically, we randomly select tokens from ground-truth boxes, compute their ERFs, and merge the results by taking the maximum value across voxel locations. 
We observe that UniTR~\cite{wang2023unitr} with a window-based transformer captures local information, limiting its coverage of larger objects and overall environmental context, while projection errors introduce additional noise. 
Vanilla Mamba provides long-range modeling and global receptive field but falls in capturing information effectively. 
The Hybrid Mamba Block improves spatial modeling, while misaligned modalities shift attention to irrelevant areas. The inclusion of HFL mitigates this misalignment.
Next, we compare various linear attention mechanisms on ERFs. Mamba demonstrates stronger coverage and focuses more effectively on key objects, whereas RWKV~\cite{peng2023rwkv} and RetNet~\cite{sun2023retentive} exhibit additional noise and fail to attend to certain crucial targets. Finally, Tab.~\ref{tab:med_linear} shows that although all linear attentions exceed the baseline, Mamba achieves the highest performance, reaffirming our choice of Mamba.

\begin{figure}[t]
    \centering
    \footnotesize
    \includegraphics[width=0.98\linewidth]{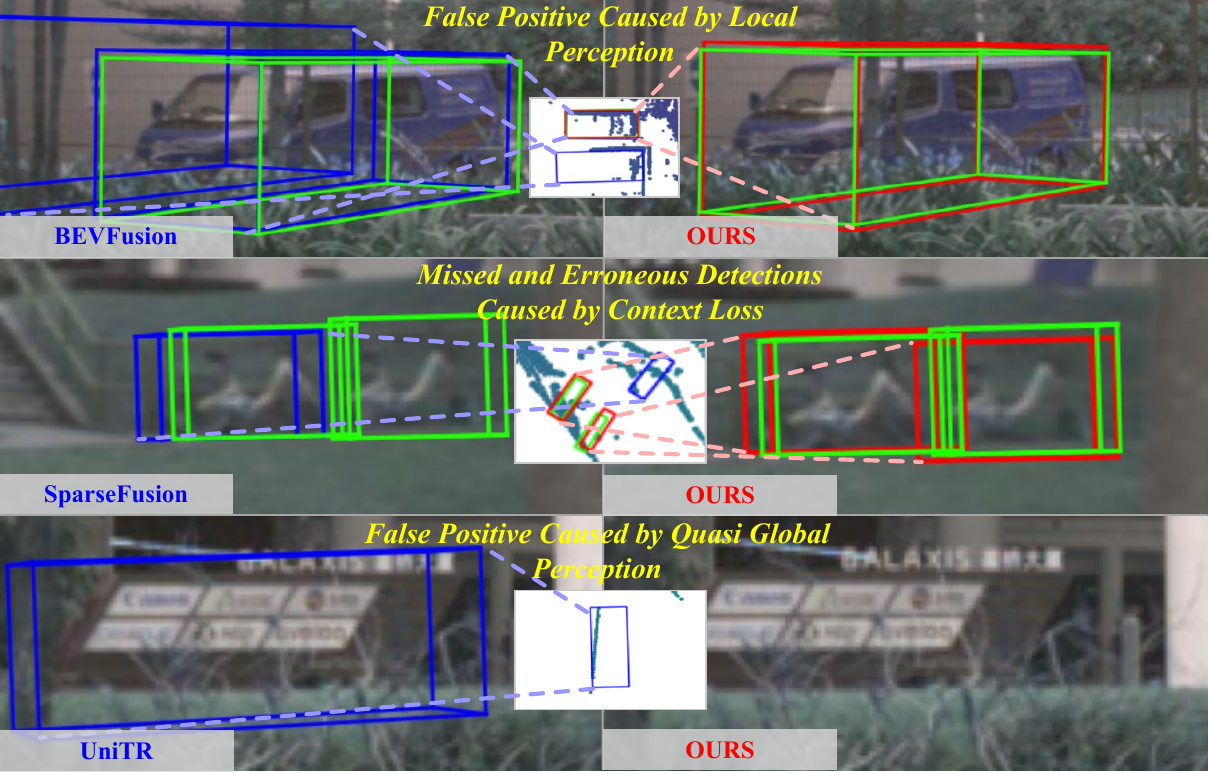}
    \vspace{-0.1cm}
    \caption{\textbf{Qualitative comparison of different fusion strategies.} The \textcolor[RGB]{33, 255, 5}{\textbf{green}}, \textcolor[RGB]{0,0,255}{\textbf{blue}} and \textcolor[RGB]{255,0,0}{\textbf{red}}
    boxes indicate ground truth, output of previous methods, and our results, respectively.}

    \label{fig:exp_denseglobal}
    \vspace{-0.5cm}
\end{figure}

\subsubsection{Visualization and Analysis.}
In this section, we present a visual analysis of three prominent fusion strategies, highlighting the limitations that our method effectively addresses. As shown in Fig.~\ref{fig:exp_denseglobal}, Dense Local Fusion (BEVFusion~\cite{liu2023bevfusion}) struggles with long-range dependency modeling. The localized receptive field and errors in projecting image features to BEV cause a single object to disperse across multiple BEV regions, resulting in false positives. Sparse Global Fusion (SparseFusion~\cite{zhou2023sparsefusion}) faces issues when background context is crucial for object differentiation. The loss of background information causes misalignment between image and LiDAR features, resulting in missed or incorrect detections. Similarly, Quasi-Dense Global Fusion (UniTR~\cite{wang2023unitr}) struggles to capture objects requiring long-range dependencies, especially those that don't fit within a single window or rely on background differentiation. This limited observation makes it difficult to accurately determine the object's category, leading to false positives.
In contrast, our method addresses these limitations by employing a Dense Global Fusion framework.

\subsection{Robustness Studies}
\begin{table}[t!]
\centering

\scalebox{0.63}{
\begin{tabular}{l|cc|cc|cc|cc}
\toprule
\multicolumn{1}{c|}{{Approach}}& \multicolumn{2}{c|}{{Clean}} & \multicolumn{2}{c|}{{Missing F}} & \multicolumn{2}{c|}{{Preserve F}} & \multicolumn{2}{c}{{Stuck}} \\
 & {mAP} & {NDS} & {mAP} & {NDS} & {mAP} & {NDS} & {mAP} & {NDS} \\
\hline
\hline
PointAugmenting \cite{wang2021pointaugmenting} & 46.9 & 55.6 & 42.4 & 53.0 & 31.6 & 46.5 & 42.1 & 52.8 \\
MVX-Net \cite{sindagi2019mvx} & 61.0 & 66.1 & 47.8 & 59.4 & 17.5 & 41.7 & 43.8 & 58.8 \\
TransFusion \cite{bai2022transfusion} & 66.9 & 70.9 & 65.3 & 70.1 & 64.4 & 69.3 & 65.9 & 70.2 \\
BEVFusion \cite{liang2022bevfusion} & 67.9 & 71.0 & 65.9 & 70.7 & 65.1 & 69.9 & 66.2 & 70.3 \\
UniTR~\cite{wang2023unitr}& 70.5 & 73.3 & 68.5 & 72.4 & 66.5 & 71.2 & 68.1 & 71.8 \\
\hline
Ours (full dataset)& 72.7 & 75.0 & 70.6 &  74.1&  67.9& 72.3 &  71.3 &  74.2 \\
\bottomrule
\end{tabular}
}
\vspace{-0.2cm}
\caption{Evaluating the robustness on degradation conditions. ``F'' is the front camera, and Stuck means unsynchronized timestamp between LiDAR and cameras.}
\label{tab:robost_cam}
\vspace{-0.25cm}
\end{table}

Following the BEVFusion~\cite{liang2022bevfusion} evaluation protocols, we evaluate our method on different degradation conditions, including missing the front camera (``Missing F'' in Tab.~\ref{tab:robost_cam}), only preserving the front camera (``Preserve F'' in Tab.~\ref{tab:robost_cam}) and unsynchronized LiDAR and camera timestamps (``Stuck'' in Tab.~\ref{tab:robost_cam}). As shown in Tab.~\ref{tab:robost_cam}, our model demonstrates superior robustness compared to other methods. Notably, our method performs particularly well in the "Stuck" setting, suggesting that global perception is advantageous in handling unsynchronized scenarios.

\begin{table}[t]
\centering
\scalebox{0.85}{
\begin{tabular}{c|c c}
\toprule
Method & DE$_{\rm avg}$ $\downarrow$ & CR$_{\rm avg}$ $\downarrow$ \\ 
\hline
\hline

FusionAD~\cite{ye2023fusionad} & 0.81 & 0.12 \\
\hline
+\textbf{MambaFusion} (ours) & 0.70 & 0.10 \\
\bottomrule
\end{tabular}
}
\vspace{-0.2cm}
\caption{End-to-end performance on nuScenes dataset.}
\label{tab:fusionAD}
\vspace{-0.6cm} 
\end{table}

\subsection{Generality: MambaFusion for End-to-end AD}
Our approach aims to achieve generalizable multi-modality long-range modeling. As shown in Tab.~\ref{tab:fusionAD}, integrating it into FusionAD~\cite{ye2023fusionad} yields improvements in motion planning, reducing average Displacement Error ($\rm  DE_{\rm avg}$) and Collision Rate ($\rm CR_{\rm avg}$). We further validate its effectiveness on Waymo~\cite{sun2020scalability} and in BEV segmentation on nuScenes (see supplementary material).  Collectively, these results underscore the robustness and versatility of our method.

\noindent\textbf{More Experiments.} We conduct experiments on different backbones and resulotions in supplementary material.
\section{Conclusion}

In this paper, we introduce the first dense global fusion framework for multi-modal 3D object detection. The core components contributing to our success are the efficient Hybrid Mamba Block and the effective height-fidelity LiDAR encoding. The former efficiently aggregates global information and the latter enables better feature alignment between modalities. Our method achieves superior performance on the competitive nuScenes benchmark, demonstrating its effectiveness in real-world scenarios. Analytical experiments validate the efficacy of our proposed approach and provide insights into the reasons behind its performance gains.
We hope that our work will inspire further research in multi-modal 3D object detection and encourage the exploration of Mamba in related fields.

{
    \small
    \bibliographystyle{ieeenat_fullname}
    \bibliography{main}
}


\end{document}